\pdfoutput=1

\documentclass[11pt]{article}

\usepackage[]{ACL2023}

\usepackage{times}
\usepackage{latexsym}

\usepackage[T1]{fontenc}

\usepackage[utf8]{inputenc}

\usepackage{microtype}

\usepackage{multirow}
\usepackage{graphicx}
\usepackage{algorithm}
\usepackage{algorithmic}
\usepackage{CJK}
\usepackage{xcolor}
\usepackage{makecell}
\usepackage{xspace}
\usepackage{newfloat}
\usepackage{listings}
\usepackage{enumitem}
\usepackage{amsmath}

\newcommand{\method}{StyleAP\xspace}
\newcommand{\dataset}{MSMT\xspace}

\title{Controlling Styles in Neural Machine Translation with Activation Prompt
}

\author{
    Yifan Wang\textsuperscript{\rm 1,2\thanks{
    *Work done while Y. Wang was an
intern at ByteDance.}}, 
    Zewei Sun\textsuperscript{\rm 2}, 
    Shanbo Cheng\textsuperscript{\rm 2}, 
    Weiguo Zheng\textsuperscript{\rm 1}, 
    Mingxuan Wang\textsuperscript{\rm 2} \\
    \textsuperscript{\rm 1} Fudan University, 
    \textsuperscript{\rm 2} ByteDance \\
    \texttt{isivan.wang@gmail.com}, ~\texttt{zhengweiguo@fudan.edu.cn} \\
    \texttt{\{sunzewei.v,chengshanbo,wangmingxuan.89\}@bytedance.com}
}

\begin{document}
\maketitle

\begin{abstract}

Controlling  styles in neural machine translation (NMT) has attracted wide attention, as it is crucial for enhancing user experience.
Earlier studies on this topic typically concentrate on regulating the level of formality and achieve some progress in this area. However, they still encounter two major challenges.
The first is the difficulty in style evaluation. The style comprises various aspects such as lexis, syntax, and others that provide abundant information. Nevertheless, only formality has been thoroughly investigated.
The second challenge involves excessive dependence on incremental adjustments, particularly when new styles are necessary. 
To address both challenges, this paper presents a new benchmark and approach. A \textbf{m}ultiway \textbf{s}tylized \textbf{m}achine \textbf{t}ranslation (\textbf{\dataset}) benchmark is introduced, incorporating diverse  categories of styles across four linguistic domains.
Then, we propose a method named
\textbf{style} \textbf{a}ctivation \textbf{p}rompt (\textbf{\method}) 
by retrieving prompts from stylized monolingual corpus, which 
does not require 
extra fine-tuning.
Experiments show that \method could effectively control the style of translation and achieve remarkable performance.
\end{abstract}

\section{Introduction}
Natural language texts can be written in various styles while preserving the content, such as politeness, formal, classical, and many others ~\cite{HOVY1987689-intro,jing2019neural,fu2018style}. 
Styles are crucial  for communication since every sentence should fit a specific scenario and the appropriate style makes it more user-centric.
A speaker needs to switch the styles of words to adapt to different conditions. 
Using the inappropriate style can be impolite or ridiculous in some societies and result in serious cultural conflicts~\cite{nida2021theory}.

As a cross-lingual generation problem,  machine translation performance heavily relies on the appropriate style. 
Therefore, many commercial translation systems provide multiple style choices, such as Portuguese (European vs. Brazilian) and English (American vs. British) in DeepL\footnote{\url{https://www.deepl.com}}, Korean (Honorific vs. Non-honorific) in Papago\footnote{\url{https://papago.naver.com}}, Chinese (Modern vs. Classical) in Volctrans\footnote{\url{https://translate.volcengine.com}}.

\begin{figure}[t]
    \centering
    \includegraphics[width=1.0\columnwidth,trim=0 30 0 20]{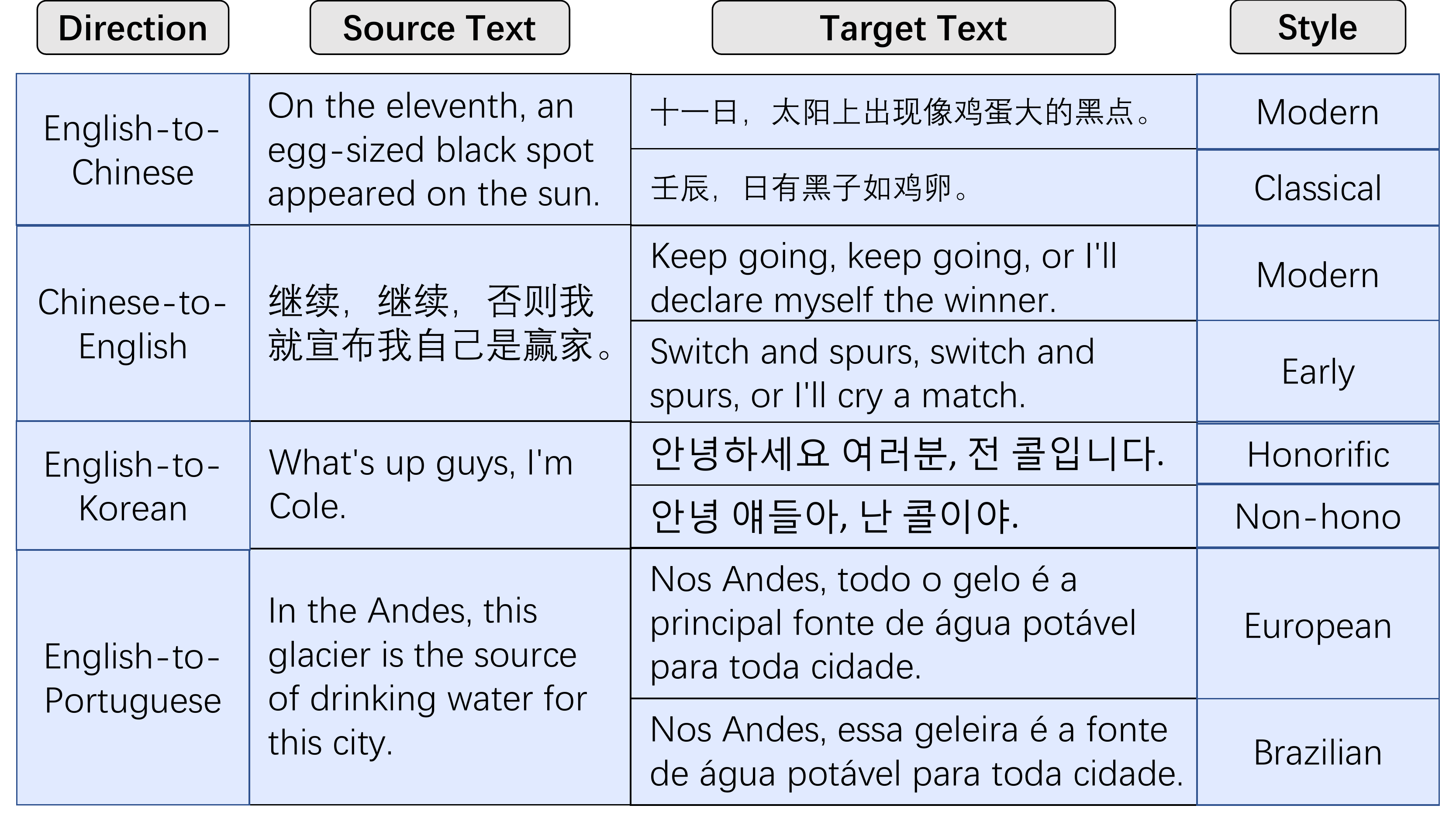} 
    \caption{Examples of stylized translation. For each language pair, two different translation styles are shown.}
    \label{introduction}
\end{figure}

Recently, controlling style in machine translation has also drawn much attention in the academic community ~\cite{yamagishi-etal-2016-controlling-voice,michel-neubig-2018-extreme-gender,feely-etal-2019-controlling-japan}.
Formally, stylized machine translation refers to translating the source sentence into different styles with certain attributes while the translation quality remains satisfactory, as the cases showed in Figure~\ref{introduction}. 
Many previous studies have explored the task and gained promising results~\cite{sennrich-etal-2016-controlling,DBLP:conf/eacl/WintnerMSRP17-personalized,wang-etal-2021-towards}. However,  challenge still remains in two aspects.

The first challenge is about the benchmark. 
The style of natural languages  consists of many aspects like word preference and grammar structure.
However, the well-studied benchmark datasets mainly focus on the formality and politeness of European languages. 
Due to this limitation, previous work restricts styles to a relative narrow scene. 
In addition, most of the test sets of the previous work have only one reference rather than multiple stylized references, which hinders the automatic evaluation for different styles.
As such, a benchmark involving more diverse styles, multiple stylized references and beyond European languages is greatly needed. 

The second challenge is about the iterative training framework. 
Most related work heavily relies on fine-tuning with new stylized data~\cite{sennrich-etal-2016-controlling,wang-etal-2021-towards}. 
Basically, they collect stylized bilingual texts and append tags before the sentence, then conduct fine-tuning to adapt the model to the given style~\cite{sennrich-etal-2016-controlling}. 
However, parallel data in specific styles is pretty sparse and costly to gather. 
Furthermore, in this way, we have to re-tune the model every time we want to add new styles, which is inconvenient.

Correspondingly, this paper contributes in terms of both benchmark and approach:

For the benchmark, we re-visit this task and push the boundary of styles to a wider range of language phenomena. We propose a dataset \textbf{\dataset}, including four directions with diverse language styles. We collect related public corpus as training sets and provide newly labeled sentences as test sets. Each source sentence has two references in different styles, which is convenient for automatic evaluation. By broadening the category and providing standard datasets, we hope to effectively push the development of this field.

For the approach, we propose \textbf{style} \textbf{a}ctivation \textbf{p}rompt (\textbf{\method}), 
a method to avoid re-tuning time after time.
The main idea is to extract one sentence of the target style as a prompt to guide the main sentence translation style.
The intuition is straightforward. We assume that once the model has been trained on all kinds of data with various styles, it has the potential to generate any style as far as correctly activated. 
We can activate the ability by language model since it tends to maintain the sequence consistency~\cite{SunWZZHCL22}. 
And the prompt can be easily retrieved in a specific stylized monolingual corpus. 
In a word, we can obtain a ``plug-and-play'' model for any new generation style with mere stylized monolingual data instead of iterative fine-tuning. The experiments show that our approach achieves explicit style transformation while well maintaining the text semantics.

\section{Related Work}



\subsection{Style Transfer for Machine Translation} Existing studies on style transfer mainly focus upon formality~\cite{feely-etal-2019-controlling-japan,DBLP:conf/ijcai/WuLLXCZH21-Dual}. They can be roughly divided into two groups: supervised methods and unsupervised methods. \citet{sennrich-etal-2016-controlling} propose side constraints to control politeness and shows that substantial improvements can be made by limiting translation to the required level of politeness. \citet{niu-etal-2017-study} propose a Formality-Sensitive Machine Translation (FSMT) scenario where lexical formality models are used to control the formality level of NMT product. Since the parallel sentences are of unknown formality level, some work focus on the unsupervised way. \citet{Niu_Carpuat_2020} introduce Online Style Inference (OSI) to generate labels via a pre-trained FSMT model. 
\citet{feely-etal-2019-controlling-japan} use heuristics to identify honorific verb forms to classify the unlabeled parallel sentences into three groups of different level formality.
\citet{wang-etal-2021-towards}
propose to use source token, embedding and output bias to control different styles and achieve a remarkable performance. \citet{DBLP:conf/aaai/WuWL20-benchmark} propose a machine
translation formality corpus.
Diverse translation is also related to this work~\cite{SunHWDC20,WuFS20}.

\subsection{Adaptive via In-Context Learning}

Recent work shows that prompting the large language models (LMs) like GPT-3~\cite{DBLP:conf/nips/BrownMRSKDNSSAA20-gpt3} with
a few examples in the context can further
leverage the inductive bias of LMs to solve different NLP tasks~\cite{Wang2022TrainingDI}.
This part of work shows the adaptive ability of LMs learned from analogy. Our work is inspired by it, but we work under the iterative training situation where the supervised data is pretty sparse. 

As prompts play a vital role in generic in-context learning, recent work propose different prompting strategies. 
\citet{DBLP:journals/corr/abs-2102-12206-domain,DBLP:journals/corr/abs-2209-11409} select the representative keywords of the field for domain adaptation. 
\citet{DBLP:journals/corr/abs-2212-10313} capture keywords of images as prompts for multimodal translation.
\citet{DBLP:conf/acl/HambardzumyanKM20-embeddings} put special tokens into the input and use continuous embeddings as prompts and \citet{DBLP:conf/acl/LiL20-prefix} directly optimize prompts in the continuous space. Besides, there is a research direction focusing on retrieval.
These methods use two main representations for generating demonstrations. As for the sparse representations, they focus on a rule-based score such as Okapi BM25~\cite{DBLP:journals/ftir/RobertsonZ09-bm25} for retrieval. \citet{Wang2022TrainingDI} use this method to improve model performance on four NLP tasks. The dense representations are generated by the pre-trained autoencoder model and have higher recall performance on most NLP tasks such as machine translation~\cite{Cai2021NeuralMT-retrieval}. For the sake of accuracy and storage, we use dense representations for retrieval in this paper.

\begin{table*}[t]\footnotesize
\centering
\begin{tabular}{c|cc|cc|cc|cc}
\Xhline{3\arrayrulewidth}
& \multicolumn{2}{c|}{en$\rightarrow$zh}     & \multicolumn{2}{c|}{zh$\rightarrow$en}     & \multicolumn{2}{c|}{en$\rightarrow$ko} & \multicolumn{2}{c}{en$\rightarrow$pt} \\ \hline
Styles & \multicolumn{1}{c}{Modern} & Classical      & \multicolumn{1}{c}{Modern} & Early       &  \multicolumn{1}{c}{Honorific}  & Non-hono  & \multicolumn{1}{c}{European} & Brazilian \\ \hline
Monolingual & \multicolumn{1}{c}{22M}   & 967K          & \multicolumn{1}{c}{22M}  & 83.2K          &  \multicolumn{1}{c}{20.5K}      & 20.5K  & \multicolumn{1}{c}{168K} & 234K \\ \hline
Parallel & \multicolumn{2}{c|}{9.12M}          & \multicolumn{2}{c|}{9.11M}         & \multicolumn{2}{c|}{271K}  & \multicolumn{2}{c}{412K} \\ 

Develepment & \multicolumn{2}{c|}{1,997}     & \multicolumn{2}{c|}{2,000}     & \multicolumn{2}{c|}{879} & \multicolumn{2}{c}{890} \\ 
Test & \multicolumn{2}{c|}{1,200}     & \multicolumn{2}{c|}{1,182}     & \multicolumn{2}{c|}{1,191} & \multicolumn{2}{c}{857} \\ \Xhline{3\arrayrulewidth}
\end{tabular}
\caption{\dataset Statistical Description. The table shows the number of training, development, and test sets.}
\label{benchmark}
\end{table*}

\section{Task Definition \& \dataset: A Multiway Stylized Translation Benchmark}
\label{sec2}

Stylized machine translation refers to the translations with certain language characteristics or styles on the basis of ensuring the quality of translation. Based upon the definition, we construct a stylized machine translation benchmark including four language directions. In each language direction, we give the illustration of various styles and provide corresponding training and test sets.

Different from traditional stylized machine translation studies, each group of our test sets is consist of one single source and multiple references in parallel. For example, for English-to-Chinese direction, for each English source sentence, we have two parallel Chinese references: classical style and modern style. In this way, we can automatically evaluate the style transformation by measuring the similarity between the stylized hypothesis and stylized references.

All the data has been publicly released
and the detailed number is in Table~\ref{benchmark}.
In this section, we will introduce our benchmark construction.

\subsection{English-to-Chinese Translation} 
There are two common styles for Chinese: \textit{Classical} and \textit{Modern}. 
Classical Chinese originated from thousands of years ago and was used in ancient China. 
Modern Chinese is the normal Chinese that is commonly used currently. 

The former is adopted on especially solemn and elegant occasions while the latter is used in daily life.
They vary in many aspects like lexis and syntax so can be regarded as two different styles.
In this direction, we aim at translating texts from English to Chinese in both styles. Specific data usage is as follows:

\begin{itemize}[leftmargin=*,itemsep=0pt,topsep=5pt]
    \item \textbf{Basic Parallel Data:} Cleaned WMT2021 corpus plus the back translation of the subset of an open source corpus containing classical Chinese and modern Chinese~\footnote{\url{https://github.com/NiuTrans/Classical-Modern}}.
    \item \textbf{Stylized Monolingual Data:} The open source corpus containing classical Chinese and modern Chinese and the Chinese part of WMT2021.
    \item \textbf{Development Set:} Newstest2019.
    \item \textbf{Test Set:} English-Classical-Modern triplet parallel data annotated by language experts.
\end{itemize}


\subsection{Chinese-to-English Translation}
There are two common styles for English: \textit{Early Modern} and \textit{Modern}.
Early Modern English in this paper refers to English used in the Renaissance such as Shakespearean plays. 
Modern English is the normal English used currently. 

The former one is mostly seen in Shakespearean play scripts like Hamlet while the latter one is used in the daily life.
They vary in many aspects like grammatical constructions such as two second person forms, thou and you. Therefore, they can be regarded as two styles.
In this direction, we aim at translating texts from Chinese to English in both styles. Specific data usage is as follows:

\begin{itemize}[leftmargin=*,itemsep=0pt,topsep=5pt]
    \item \textbf{Basic Parallel Data:} Cleaned WMT2021 corpus plus the back translation of a crawled corpus: \textit{The Complete Works of William Shakespeare}~\footnote{\url{http://shakespeare.mit.edu/}}.
    \item \textbf{Stylized Monolingual Data:} An open source dataset\footnote{\url{https://github.com/harsh19/Shakespearizing-Modern-English}} containing early modern and modern English and the English part of WMT2021.
    \item \textbf{Development Set:} Newstest2019.
    \item \textbf{Test Set:} Chinese-Early-Modern triplet parallel data annotated by language experts.
\end{itemize}


\subsection{English-to-Korean Translation}
There are seven verb paradigms or levels of verbs in Korean, each with its own unique set of verb endings used to denote the formality of a situation. We simplify the classification and roughly divide them into two groups: \textit{Honorific} and \textit{Non-honorific}

The former is used to indicate the hierarchical relationship with the addressee such as from the young to the old, from the junior to the senior.
The latter is used in daily conversations between friends.
They vary in some lexical rules so can be regarded as two styles.
In this direction, we aim at translating texts from English to Korean in both styles. Specific data usage is as follows:

\begin{itemize}[leftmargin=*,itemsep=0pt,topsep=5pt]
    \item \textbf{Basic Parallel Data:} IWSLT2017~\footnote{\url{https://wit3.fbk.eu/}} plus the back translation of an open source dataset~\footnote{\url{https://github.com/ezez-refer/Korean-Honorific-Translation}} containing honorific and non-honorific.
    \item \textbf{Stylized Monolingual Data:} The open source dataset and the crawled corpus from a public translation tool\footnote{\url{https://papago.naver.com/}}.
    \item \textbf{Development Set:} IWSLT17.
    \item \textbf{Test Set:} English-Honorific-Non-honorific triplet parallel data annotated by language experts language experts.
\end{itemize}


\begin{figure*}[t]
\centering
\includegraphics[width=0.9\columnwidth,trim=200 20 200 20]{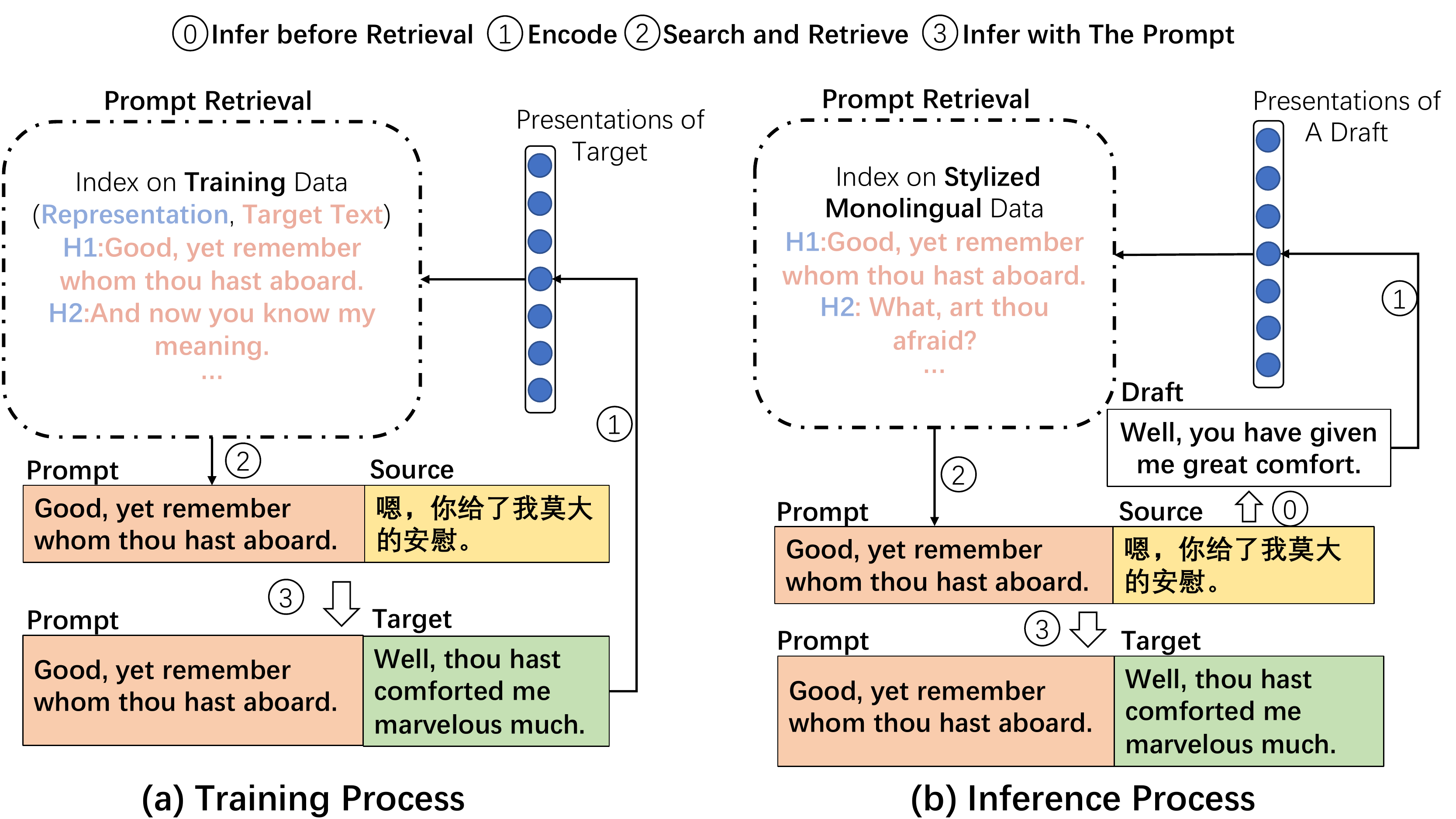} 
\caption{The proposed model training and inference process. We provide an example and mark the entire retrieval process in order. During training process, it includes encoding, retrieving and predicting, while it has an extra predicting operation during inference. 
}
\label{method}
\end{figure*}

\subsection{English-to-Portuguese Translation}

There are two common styles for Portuguese: \textit{European} and \textit{Brazilian}. 
European Portuguese is mostly used in Portugal.
Brazilian Portuguese is mostly used in Brazil.


They vary in some detailed aspects like pronunciation, grammar and spelling, so can be regarded as two different styles.
In this direction, we aim at translating texts from English to Portuguese in both styles. Specific data usage is as follows:

\begin{itemize}[leftmargin=*,itemsep=0pt,topsep=5pt]
    \item \textbf{Basic Parallel Data:} IWSLT2017.
    \item \textbf{Stylized Monolingual Data:} European \& Brazilian part of the parallel data.  
    \item \textbf{Development Set:} IWSLT17.
    \item \textbf{Test Set:} English-European-Brazilian triplet parallel data annotated by language experts.
\end{itemize}


\subsection{Evaluation}
Previous style evaluation relies on human resources, which is costly and slow. Since our test sets are all multiway, we can evaluate our stylized hypothesis with the corresponding reference to take both quality and style into consideration at a small cost. Moreover, human evaluation is inevitably subjective while our test sets can guarantee the comparison stability.

\section{Style Activation Prompt}

Prior work usually uses the fixed tag to control the generation with expected attributes~\cite{sennrich-etal-2016-controlling}. However, tag-based methods rely on a large amount of labeled parallel data, requiring relabeling and retraining of models when new styles need to be generated. 

We go back to a standard NMT model. During the generation process, the NMT model tries to maximize the probability of the generation sentence. When predicting the $i$-th token, the model searches from the vocabulary to maximize:

\begin{equation}
    P(y_i|x, y_{j<i}) \nonumber
\end{equation}

where $y_{j<i}$ means the past words, indicating that the previous inference results can affect the subsequent generation. Therefore, our intuition is to control the generation style by taking advantage of stylized language model. We suggest that once the basic model has been trained on the data in various kinds of styles, we can activate the ability with the contextual influence.

Specifically, we retrieve an instance as a prompt from the stylized corpus and use it to instruct the NMT model to generate the sentence with the same attributes. 
To adapt to the prompt training, we extract every sentence in the basic parallel data and retrieve one similar sentence as the prompt.
The whole framework is in Figure~\ref{method}.
We introduce the details of our proposed method as follows. 

\subsection{Prompt Retrieval}
The prompt retrieval procedure aims at finding the proper sentence prompt. First, we construct a candidate datastore $D$ that contains many $(r, y)$ key-value pairs, where $r$ is the representation of $y$. In this paper, we use a multilingual pre-trained language model XLM-R~\cite{DBLP:journals/corr/abs-1911-02116-xlmr} to obtain the sentence representation.
By calculating the similarity between the query representation $h$ and keys, we can extract the needed sentence $y$:
$$y=\arg\min\limits_{r\in\mathcal{D}}Distance(h, r)$$
where the search tool is Faiss\cite{johnson2019billion-faiss}, a library for efficient similarity search and clustering of dense vectors. 

\subsection{Training}
In the training stage, the goal is to retrieve a similar prompt with the current sentence to make the model adapt to the inference pattern.
Specifically, we iterate each target-side sentence in the basic parallel data as a query and retrieve the most similar sentence as its prompt.

After obtaining the prompt, we concatenate the prompt and the query sentence by a special token as: 
$$prompt, [s], src \rightarrow prompt, [s], trg$$
We train the model with this kind of data and normal data together to learn the prompt-based generation as well as basic translation.

\begin{table*}[t]\footnotesize
\centering
\begin{tabular}{c|cc|cc|cc|cc|c}
\Xhline{3\arrayrulewidth}
& \multicolumn{2}{c|}{en$\rightarrow$ zh}     & \multicolumn{2}{c|}{zh$\rightarrow$en}     & \multicolumn{2}{c|}{en$\rightarrow$ko} & \multicolumn{2}{c|}{en$\rightarrow$pt} &\\ \hline
Styles & \multicolumn{1}{c}{Modern} & Classical      & \multicolumn{1}{c}{Modern} & Early       & \multicolumn{1}{c}{Honorific}  & Non-hono  & \multicolumn{1}{c}{European} & Brazilian & Average \\ \hline
Baseline & \multicolumn{1}{c}{25.00}   & 13.86          & \multicolumn{1}{c}{26.73}  & 14.28          & \multicolumn{1}{c}{20.65}      & 17.48  & \multicolumn{1}{c}{31.30} & 32.86 & 22.77\\ 
Transfer & \multicolumn{1}{c}{24.87}   &   20.88   & \multicolumn{1}{c}{11.05}  &    7.46      & \multicolumn{1}{c}{\textless5} & \textless5 & \multicolumn{1}{c}{32.84} & 32.59 & \textless20 \\ 
Tag-tuning & \multicolumn{1}{c}{28.43}  & 21.21 & \multicolumn{1}{c}{\textbf{27.16}}  & 14.48 & \multicolumn{1}{c}{21.05} & \textbf{21.11} & \multicolumn{1}{c}{33.67}  & 33.84  & 25.11\\
\method & \multicolumn{1}{c}{\textbf{29.73}}  &  \textbf{24.98}   & \multicolumn{1}{c}{26.76}   &  \textbf{17.72} & \multicolumn{1}{c}{\textbf{21.65}} &   20.67   & \multicolumn{1}{c}{\textbf{33.82}}  & \textbf{34.27} & \textbf{26.20}\\ \Xhline{3\arrayrulewidth}
\end{tabular}
\caption{BLEU results on the multiple stylized references. The experiment of en$\rightarrow$ko Translation Transfer fails and yields non-sense results due to the data scarcity. Overall, \method achieves the best results.}
\label{result}
\end{table*}

\subsection{Inference}
In the inference stage, we first translate the source sentence roughly. Then the draft hypothesis is used as the query. The candidate datastore is constructed with the monolingual data in the given style. After retrieving the prompt, we append it to the beginning of the source sentence with the special token. After the second inference, the hypothesis can be obtained by splitting the token.

\subsection{Advantages}
We conclude the advantages of \method as follows:
\begin{itemize}[leftmargin=*,itemsep=0pt,topsep=5pt]
    \item \method does not need any architecture modification and is easy to deploy. 
    \item \method does not need to assign various tags to all kinds of styles.
    \item Extra tuning is no longer needed when it comes to a new style.  We only need to retrieve the prompt from the new monolingual stylized corpus and then generate the given style.
\end{itemize}

\section{Experiments}
In this section, we will introduce the details of our experiments. 

\subsection{Setup}

We first compare our method with other baseline models on four tasks. 
Then, we design a manual evaluation to assess whether our method maintains translation quality and achieves diversity. 
All experiments are implemented in the following settings. 

\subsubsection{Data \& Preprocessing}

In the previous section, we introduce our provided stylized NMT benchmark \dataset. Our designed experiments and analysis are based upon this benchmark. The statistics information of this benchmark is shown in the Table \ref{benchmark}. 

We use SentencePiece\cite{kudo-richardson-2018-sentencepiece} to jointly learn an unsupervised tokenizer. We preprocess the training data and filter the parallel sentences with length greater than 256. We set hyper-parameters min frequency 5 and max vocabulary 32k. 

\subsubsection{Implementation Details}

Here, we introduce more details of our experiment settings. Our experiment is implemented on the open source Seq2Seq tool Neurst\footnote{\url{https://github.com/bytedance/neurst}}~\cite{DBLP:conf/acl/ZhaoWDYL21-neurst}. Our seq2seq model uses a transformer-base structure with 6 encoder layers and 6 decoder layers, attention with a layer size of 512, and word representations of size 512. We apply post-layer normalization~\cite{2016arXiv160706450L-LN}, adding dropout to embeddings and attention layers with a dropout rate of 0.1. We tie the source and target embeddings. The main training parameters are as follows. We use Adam Optimizer with $\beta_1=$ 0.9 and $\beta_2=$ 0.98. We use label smoothed cross entropy as a criterion with a label smoothing rate of 0.1. We set batch size per GPU 4096 and batch by tokens. And, we use four A100 GPUs to train our model from scratch. We save checkpoints every 1000 steps and stop training when there is no improvement in the continuous 50 checkpoints.

\subsubsection{Comparing Systems} 
We use sacreBLEU~\cite{post-2018-sacrebleu} as our metrics and compare \method with three common systems:

\begin{itemize}
    \item \textbf{Baseline:} Transformer that is trained on the raw parallel data.
    \item \textbf{Transfer:} A two-phases processing: Translate first and conduct style transfer~\cite{Syed_Verma_Srinivasan_Natarajan_Varma_2020}. We train the translation model with normal parallel data and train the transfer model with stylized data.
    \item \textbf{Tag-tuning:} A tag-based model which is generally used in other work~\cite{sennrich-etal-2016-controlling}. They add a special token as the tag at the start of the source text with the known style. In this way, the model can generate different styles with different tokens. This method needs explicit extra fine-tuning.
\end{itemize}

\subsection{Results} 
As is shown in Table~\ref{result}, we calculate the BLEU score on the test set to compare \method with the mentioned baselines. 
The Transfer methods have many drawbacks. Not only does it need two-phases training, but also yields poor results. The attempt in English-to-Korean direction even fails. Tag-tuning gains some improvements and even achieves the best performance in some directions. But overall, \method obtains the best results and outperform the other methods. At the same time, \method needs no extra tag or extra tuning when it comes to new styles. After acquiring the ability to translate with style prompt, \method can handle various styles. 

\subsection{Human Evaluation}

We also design a human evaluation to manually check the style transfer ratio as well as the translation quality preservation during the transfer.
The quality score ranges from 0 to 4. 
The style transfer ratio means the percentage of the hypothesis that meets the required style, ranging from 0 to 100.
Refer to the appendix for the specific scoring criteria and rules.

\subsubsection{Quality Preservation}

The quality results are in Table~\ref{quality}. \method achieves a comparable performance with the baseline model, which means little semantic loss within the stylized translation.

\begin{table}[t]\footnotesize
\centering
\begin{tabular}{c|cccccccc}
\Xhline{3\arrayrulewidth}
\multicolumn{1}{l|}{} & \multicolumn{2}{c|}{en$\rightarrow$zh}                          & \multicolumn{2}{c|}{zh$\rightarrow$en}                          & \multicolumn{2}{c|}{en$\rightarrow$ko}                    & \multicolumn{2}{c}{en$\rightarrow$pt}  \\ \hline
Styles                       & \multicolumn{1}{c}{M} & \multicolumn{1}{c|}{C}      & \multicolumn{1}{c}{M} & \multicolumn{1}{c|}{E}       & \multicolumn{1}{c}{H} & \multicolumn{1}{c|}{N} & \multicolumn{1}{c}{E}                           & B     \\ \hline
Base               & \multicolumn{1}{c}{3.5}   & \multicolumn{1}{c|}{\textbf{3.6}}          & \multicolumn{1}{c}{\textbf{3.6}}  & \multicolumn{1}{c|}{\textbf{3.7}}          & \multicolumn{1}{c}{1.9}     & \multicolumn{1}{c|}{1.9} & \multicolumn{1}{c}{2.7}     &   2.8 \\ 
\method     & \multicolumn{1}{c}{\textbf{3.6}}  & \multicolumn{1}{c|}{\textbf{3.6}} & \multicolumn{1}{c}{3.5}  & \multicolumn{1}{c|}{3.6} & \multicolumn{1}{c}{\textbf{2.0}}      & \multicolumn{1}{c|}{\textbf{2.0}} & \multicolumn{1}{c}{\textbf{2.9}} & {\textbf{2.9}} \\ \Xhline{3\arrayrulewidth}
\end{tabular}%
\caption{The quality of all systems, ranging from 0 to 4. \method maintains a comparable quality even the style is transformed.}
\label{quality}
\end{table}

\begin{figure}[t]
\centering
\includegraphics[width=0.9\columnwidth]{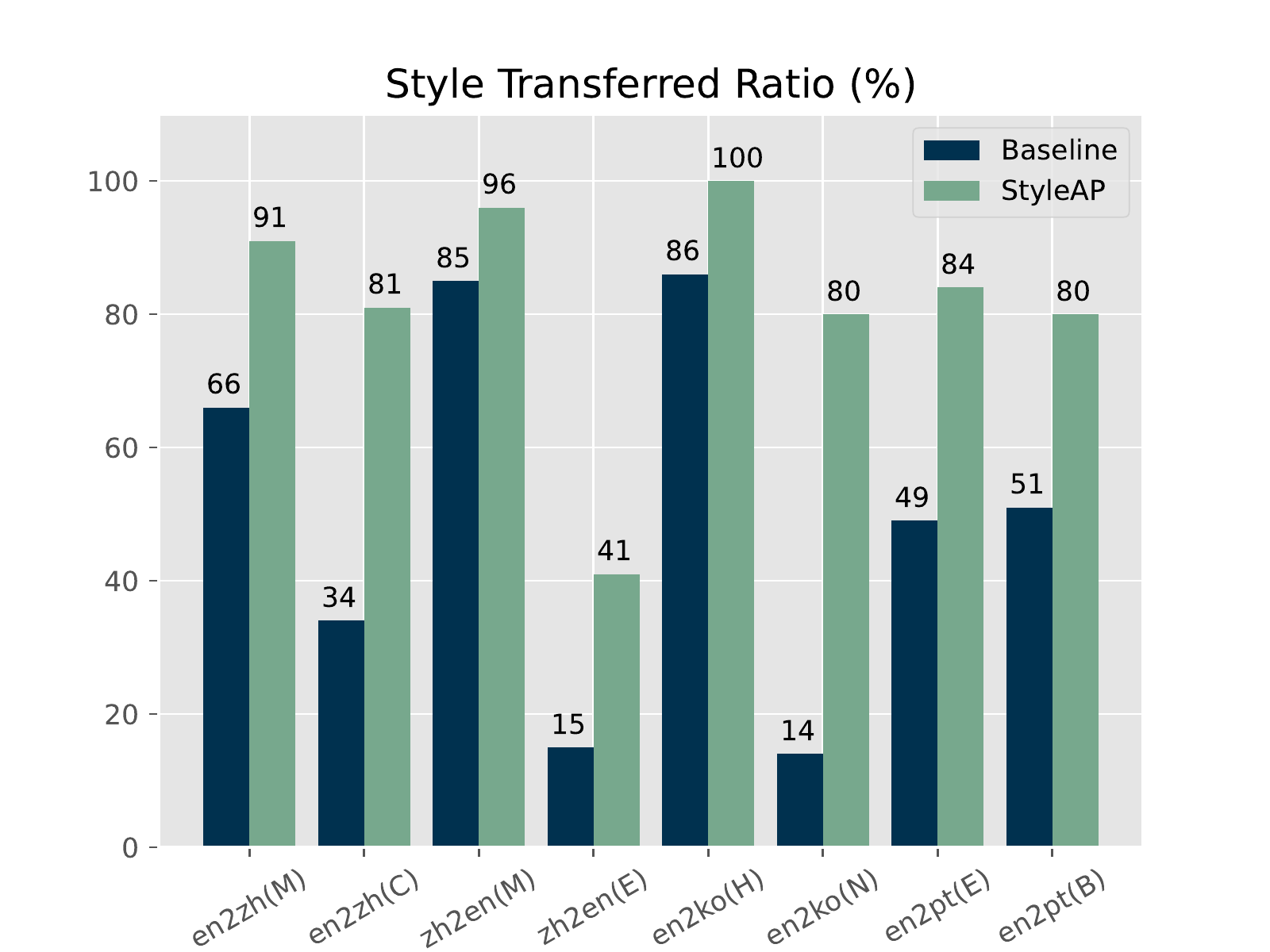} 
\caption{The percentage of successfully transferred sentences, ranging from 0 to 100. \method significantly enhances the generation ratio of certain styles.}
\label{style_ratio}
\end{figure}

\subsubsection{Style Transfer Ratio}

As is shown in Figure~\ref{style_ratio}, \method significantly enhances the transfer ratio. The only unsatisfactory is the Chinese-to-English Translation in the Early Modern style. The reason is that many sentences in the test set are very short like ``What about that?'' vs ``What of that?''. The styles for this kind of extremely short sentences are not meaningful.


\subsubsection{Conclusion} The quality scores are comparable with the baseline model, while the transferred ratios are much higher than the baseline model in the four tasks. That indicates that our method can effectively translate the source text into a sentence with specific attributes without quality loss.

\section{Analysis}
\subsection{Retrieval Strategy Matters}
\label{sec:retrieval}

There are many retrieval methods to select a similar sentence from stylized monolingual sentences. 
We conduct a detailed comparison in English-to-Chinese inference with the following strategies:
\begin{itemize}
    \item \textbf{Source:} Directly use the source text representation generated from the pre-trained multilingual language model.
    \item \textbf{Random:} Randomly choose a prompt from candidates.
    \item \textbf{Fixed:} Use the same prompt for all samples.
\end{itemize}

The results are shown in Table~\ref{retrieval}. We can see that the our strategy performs the best and the other retrieval methods face different levels of the BLEU loss. The retrieval strategy plays an important role in the translation.

\subsection{Even Unsupervised Prompts Works}
In the training phase, we assume that the retrieval range lies within the specific styles. However, one condition that is more close to the real world is that we need to retrieve prompts from more general data, which may cause the style mismatch between the sentence and the prompt. Therefore, we also conduct a unsupervised prompt retrieval in training for English-to-Chinese direction. 

As is shown in Table~\ref{unsupervised}, the unsupervised version of \method slightly drops in terms of BLEU but still outperforms Tag-tuning. It is worth mentioning that we have none of the style label of parallel data in this setting. General parallel data and monolingual stylized data is all you need. This again shows the universality and robustness of \method.

\begin{table}[t]
\centering
\begin{tabular}{l|cc}
\Xhline{3\arrayrulewidth}
        & Modern & Classical \\ \hline
\method    & 29.73  & 24.98    \\ 
-Source    & 25.51  & 18.07     \\ 
-Random & 24.72  & 15.65     \\ 
-Fixed    & 24.52  & 13.75     \\ \Xhline{3\arrayrulewidth}
\end{tabular}
\caption{Our retrieval strategy achieves the best results of BLEU.}
\label{retrieval}
\end{table}

\begin{table}[t]
\centering
\begin{tabular}{l|cc}
\Xhline{3\arrayrulewidth}
        & Modern & Classical \\ \hline
\method    & 29.73  & 24.98     \\ 
\method(U)    & 28.72  & 23.76     \\ 
Tag-tuning    & 28.43  & 21.21     \\ \Xhline{3\arrayrulewidth}
\end{tabular}
\caption{Unsupervised \method still gains the better performance than Tag-tuning.}
\label{unsupervised}
\end{table}

\begin{table*}[]
\centering
\begin{tabular}{l|l}
\Xhline{3\arrayrulewidth}
\multirow{2}{*}{Source} & \begin{CJK*}{UTF8}{gbsn}现在，\quad亲爱的奶妈，\quad哦上帝，\quad你\quad为什么\quad看起来\quad这么伤心？\end{CJK*} \\ \cline{2-2} 
                        & (Now) (good sweet Nurse) (Oh Lord) \quad(you) (why) \quad(look)  \quad(so sad)       \\ \hline
                                                                    \hline
Ref (E)                & Now, good sweet Nurse, \textbf{O} Lord, why \textbf{look'st} \textbf{thou} sad? \\ \hline
Baseline                & Now, good sweet Nurse, Oh Lord, why do you look so sad? \\ \hline
Prompt               & What say'st \textbf{thou}, my dear nurse? \\ \hline
Tagged                  & Now, dear nurse, \textbf{O} God, why look you so sad? \\ \hline
\method                     & Now, sweet nurse, \textbf{O} God, why \textbf{dost} \textbf{thou} look so sad? \\ \Xhline{3\arrayrulewidth}
\end{tabular}
\caption{An example of using \method 
to translate Chinese sentences to the Early Modern English style. The first two rows are the Chinese sentence and corresponding English 
translation of each Chinese word, respectively. } 
\label{case-table}
\end{table*}


\begin{figure}[t]
\centering
\includegraphics[width=1.0\columnwidth]{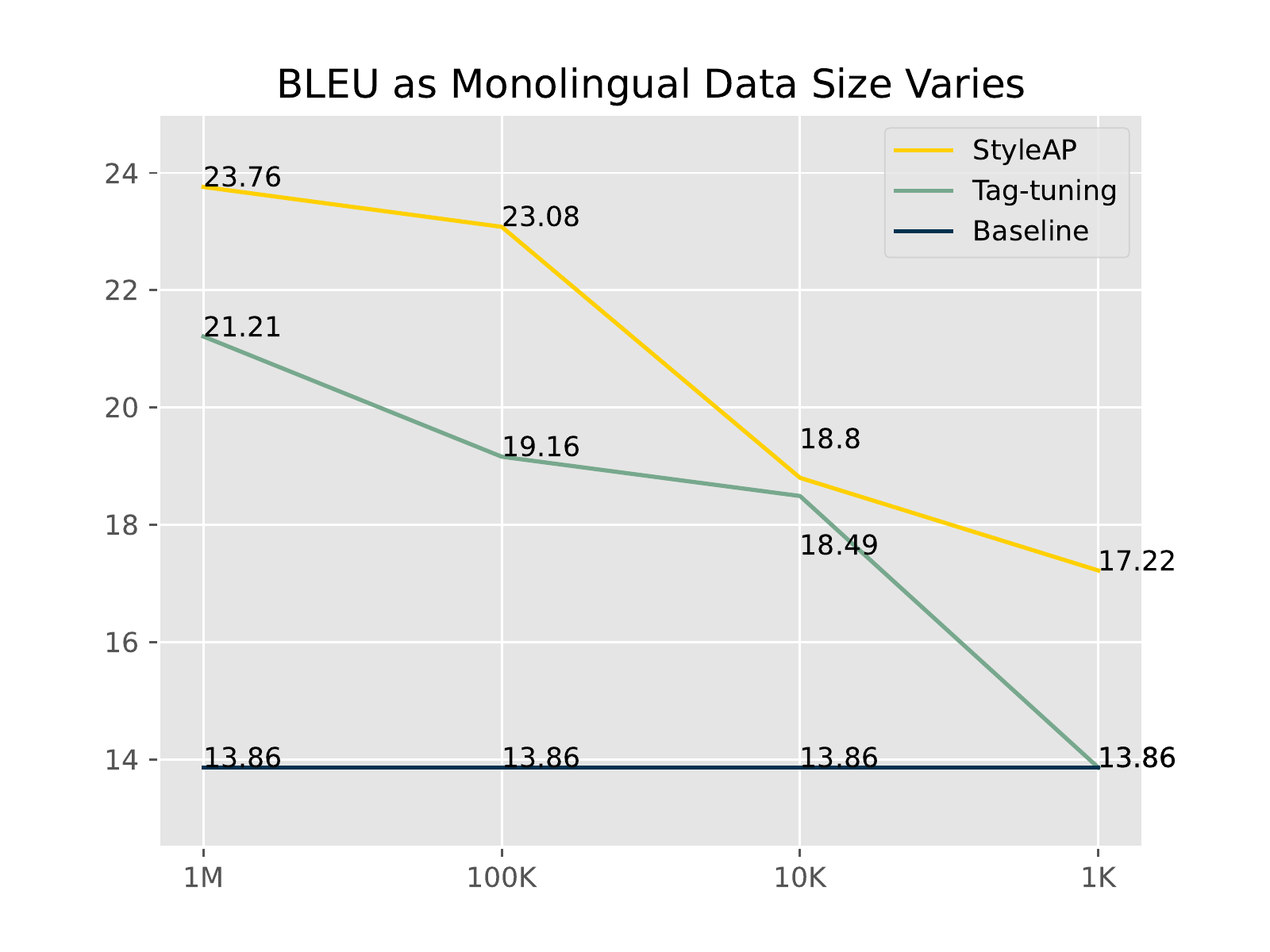} 
\caption{The impact of the size of stylized corpus on the en$\rightarrow$zh task. \method shows a consistent performance across all sizes, even the extremely few one.}
\label{size}
\end{figure}

\subsection{Consistent Performance across Sizes}

We are also interested in the situation of unbalanced data and even few stylized data. 
We implement a comparative experiment of stylized monolingual data on the en$\rightarrow$zh task.
We control the amount of labeled data at four levels: 1 million, 100 thousand, 10 thousand, and 1 thousand.
For the tag-based method, we train the tag-based model from scratch at each level. 
For a fair comparison, we use the same stylized labeled data as the tag-based method but only as the target monolingual part for retrieval. 

The results are shown in Figure~\ref{size}
, where the horizontal axis represents the sample size, and the vertical axis represents the BLEU score. For the classical direction, our method performs better in all situations. Even when we only use 1,000 labeled stylized monolingual sentences, there is still an improvement compared with the baseline model. On the contrary, the tag-based method performs poorly in few data and even has a lower BLEU score than the baseline model.

In conclusion, our method has an overall better performance than the tag-based method at different levels. Especially for extremely few samples, our method still gains significant improvements.

\subsection{Attention Score Interprets the Effect}

\begin{figure}[t]
    \centering
    \includegraphics[width=0.49\columnwidth]{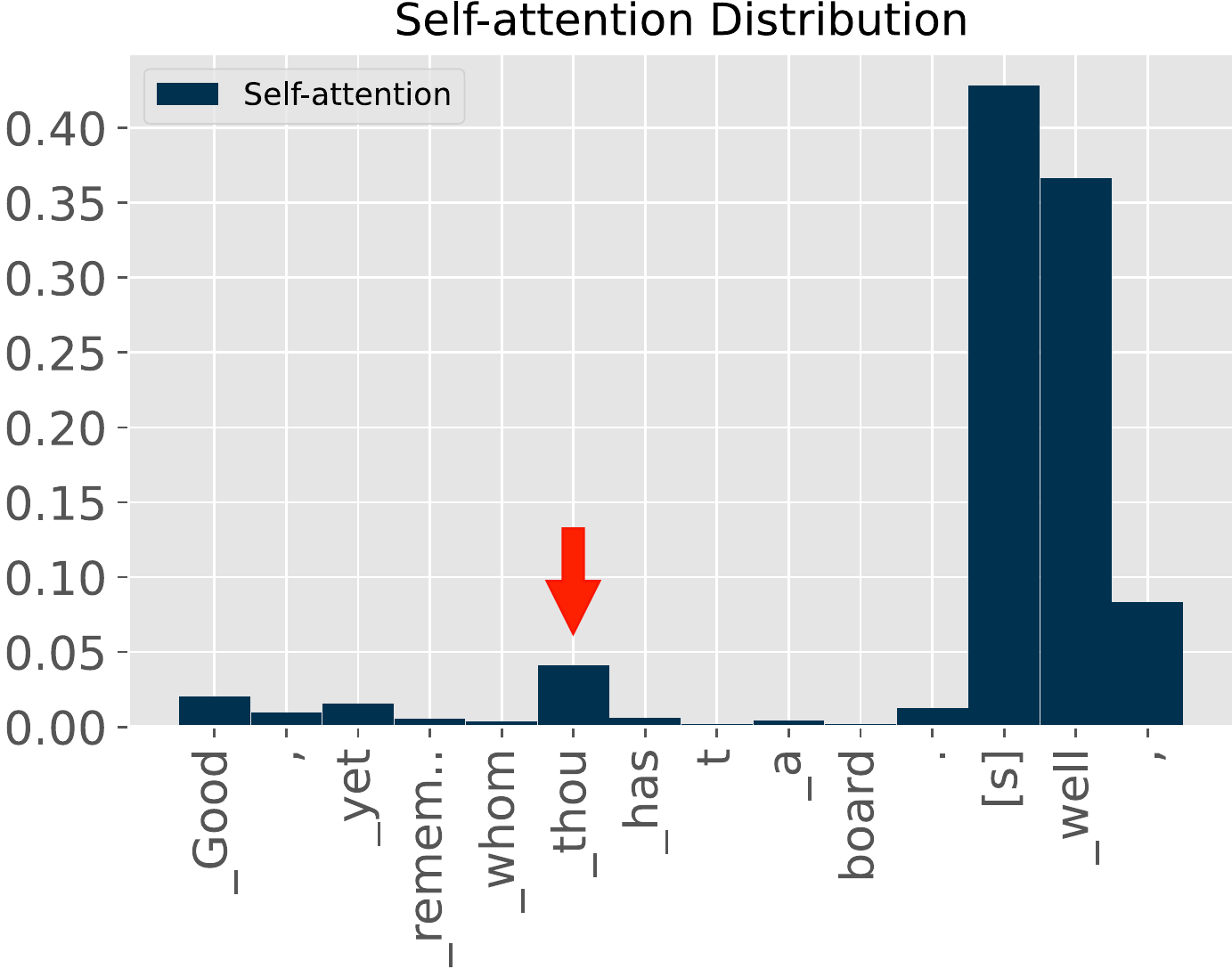} 
    \includegraphics[width=0.49\columnwidth]{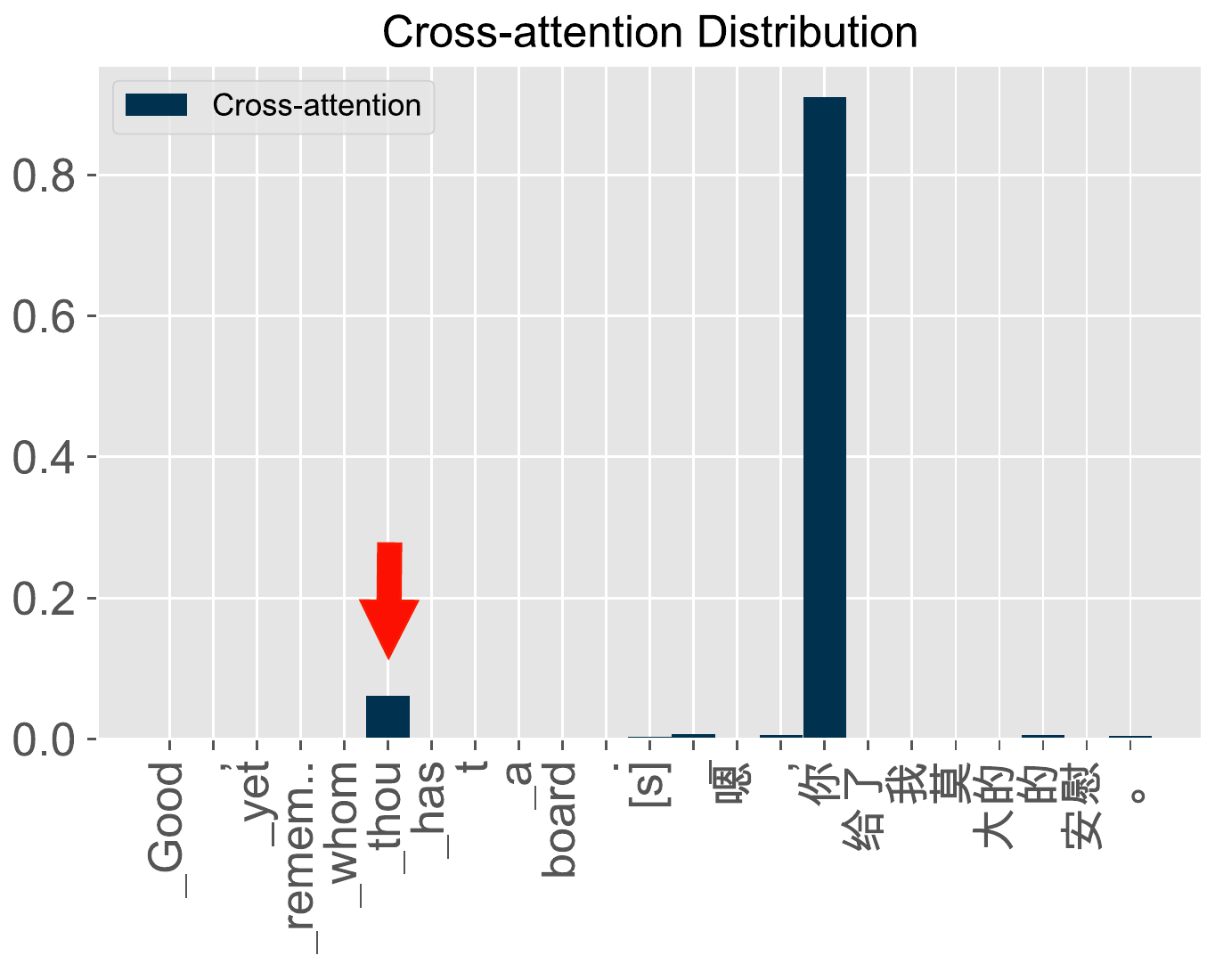}
    \
    \caption{Attention score histograms, left for self-attention, right for cross-attention. The sample comes from the hypothesis ``Well, thou hast given me great comfort.'' when predicting the token ``\_thou'' by our model. The figure shows the style effect with the attention mechanism.
    }
    \label{att}
\end{figure}

We are also interested in how the retrieved prompt affects the translation style.
Here is an example of Chinese-to-English task in Figure~\ref{att}.
The model is translating a Chinese sentence meaning ``\textit{Yeah, you have given me great comfort.}'' into English and the next token is ``thou'', which means ``you'' in early modern English.

We show the average attention scores in the Transformer decoder, left for self-attention and right for cross-attention.
For self-attention, except for some adjacent tokens, the model mainly pays attention to the token ``thou'' in the prompt which corresponds to the generating token. For the cross-attention, the model concentrates on the corresponding Chinese token 
``Ni''
(means ``you'' in Chinese) and the token ``thou'' in the prompt again.

This result suggests that our retrieval prompt could affect the generation process through the attention mechanism.

\subsection{Case Study}

Finally, one stylized Chinese-to-English translation example is listed in Table~\ref{case-table} to show the effectiveness of our method more intuitively.
As in these examples, the Chinese word ``Ni'' (``you'' in English) is translated into ``you'' in Baseline and the Tag-tuning method. However, under the guidance of the prompt which uses the early modern English word ``thou'', \method translates the word into ``thou'' correspondingly. Obviously, \method can explicitly affect the translation style with prompts.

\section{Conclusions}
In cross-lingual generation fields, most studies focus on the translation quality but ignore the style issue, which happens to be important in the real-world communication. However, the previous studies face two major challenges including the benchmark as well as the approach. For those purposes, we re-visit this task and propose a standard stylized NMT benchmark \dataset with four well-defined tasks to push the boundary of this field. We also propose a new translation style controlling method with activation prompt. With stylized prompts that are retrieved from the stylized monolingual corpus, we successfully guide the translation generation style without iterative fine-tuning. Through automatic evaluation and human evaluation, our method achieves a remarkable improvement over baselines and other methods.
A series of analysis also show the advantages of our method.

\section*{Limitation}
One limitation of \method is that one extra inference is needed for retrieval. It is mainly due to the monolingual retrieval accuracy is higher than that of crosslingual retrieval (refer to Section~\ref{sec:retrieval}). In the future, we will try stronger multilingual model to mitigate this effect.

\section*{Acknowledgments}

We would like to thank the anonymous reviewers for their constructive comments. Weiguo Zheng is the corresponding author.

\bibliography{custom}
\bibliographystyle{acl_natbib}

\clearpage
\appendix

\section{Appendix}
In this section, we supplement human evaluation criterion in this paper and more stylized translation cases. Table~\ref{tab:criterion} fully illustrates the score standard of our language experts. Table~\ref{case} shows more English examples of stylized translation.

\begin{table*}[t] \footnotesize
\centering
\begin{tabular}{c|p{1.5cm}|p{9cm}|p{3cm}}
\Xhline{3\arrayrulewidth}
\multicolumn{1}{l|}{} & Accuracy & Criterion & Description \\ \hline
\multirow{5}{*}{Quality} & 4 & The translation faithfully reflects the semantics and the translation is fluent. & There is no errors and no modification required. \\ \cline{2-4} 
 & 3 & The translated text basically reflects the semantics of the original text and is basically fluent(the subject, predicate, object and other grammatical components are in correct order), but there are a few non-keywords that are improperly used or inappropriately matched, etc. There are slight mistakes, which will not affect the understanding of the original text such as improper use of words, punctuation, capitalization, irregular date format, etc. & The meaning is basically correct, but there are partial errors, which will cause certain difficulties in understanding. \\ \cline{2-4} 
 & 2 & The translation can reflect the semantics of the original text, the translation has one or more general errors, and the translation is basically fluent (the order of grammatical components such as subject, predicate and object is correct), but there are key words that express the semantics of improper translation, omission or mistranslation of non-keywords, etc. & The meaning is basically correct, but there are partial errors, which will cause certain difficulties in understanding. \\ \cline{2-4} 
 & 1 & The translated text cannot reflect the semantics of the original text, and there are multiple serious translation errors in the translation text. One of the following situations exists: a) The translation text contains the main components of the original text, but fails to form a fluent composition due to sequence problems, logical errors, serious grammatical errors (including tenses), etc. The translation; b) The translation is basically fluent, but there are translation errors such as negation and double negation, serious omission of translation, mistranslation of keywords, and more translation of content that is not in the original text. & There are serious errors that have a greater impact on understanding. \\ \cline{2-4} 
 & 0 & The translated text cannot express the meaning of the original text at all: a) The translation text is obscure and difficult to understand, and the content expressed in the original text cannot be judged by the translation text; b) A string of repeated words and garbled characters appear; totally different/completely unrelated; d) the entire sentence is not translated. & The translated text is almost completely wrong or completely incomprehensible.  \\ \hline
\multicolumn{1}{l|}{\multirow{2}{*}{Style}} & 1 & The translation has the corresponding style. &  \\ \cline{2-4} 
\multicolumn{1}{l|}{} & 0 & The translation has not the corresponding style. &  \\ \Xhline{3\arrayrulewidth}
\end{tabular}
\caption{Human evaluation criterion.}
\label{tab:criterion}
\end{table*}

\begin{table*}[t]\footnotesize
\centering
\begin{tabular}{p{3.4cm}|p{3.4cm}|p{3.4cm}|p{3.4cm}}
\Xhline{3\arrayrulewidth}
Modern English & Baseline & Retrieved Prompt & \method \\ \hline
I swear to you, You have a good heart, and believe me, I'll tell her that. & I swear to you, you are kind, trust me, and I'll tell her. & I tell \textbf{thee}, I, that \textbf{thou} hast marred her gown. & I swear to \textbf{thee}, \textbf{thou} art kind, and believe me, I'll tell her. \\ \hline
Now I'll tell you so you don't have to ask. & Now I tell you, so you need not ask. & To tell \textbf{thee} \textbf{thou} shalt see me at Philippi. & Now I tell \textbf{thee}, so \textbf{thou} shalt not ask. \\ \hline
You're not paying attention to me. & You did not notice me at all. & God \textbf{mark} \textbf{thee} to his grace! & \textbf{Thou} \textbf{dost} not mark me. \\ \hline
If you were ever yourself, and this sadness was yours, you and your sadness were all for Rosaline. & If you were once yourself, This sorrow is yours, and both you and your sorrow are for Rosalin. & If e'er \textbf{thou} wast \textbf{thyself} and these woes thine, \textbf{thou} and these woes were all for Rosaline. & If \textbf{thou} werest thyself, This sorrow was \textbf{thy}, \textbf{Thou} and \textbf{thy} sorrow were all for Rosaline. \\ \hline
Therefore, the fact that you're awake this early tells me you've been upset with some anxiety. & Therefore you wake so early, which tells me you are uneasy about some anxiety. & Unless \textbf{thou} \textbf{tell'st} me where \textbf{thou} hadst this ring, \textbf{Thou} diest within this hour. & Therefore, \textbf{thou} \textbf{awaken'st} so early, That tells me \textbf{thou} art uneasy with some anxiety. \\
\Xhline{3\arrayrulewidth}
\end{tabular}
\caption{Examples from test sets and the results of the baseline and \method.}
\label{case}
\end{table*}

\end{document}